\documentclass[a4paper, 10pt, conference]{ieeeconf}      
\usepackage{FG2017}
\usepackage{times}
\usepackage{epsfig}
\usepackage{graphicx}
\usepackage{amsmath}
\usepackage{amssymb}
\usepackage{epstopdf}
\usepackage{multirow}
\usepackage{caption}
\FGfinalcopy 
\graphicspath{{figs/}}
\DeclareMathOperator*{\argmin}{arg\,min}

\IEEEoverridecommandlockouts                              
\title{\LARGE \bf
An All-In-One Convolutional Neural Network for Face Analysis
}

\author{\parbox{16cm}{\centering
    {\large Rajeev Ranjan, Swami Sankaranarayanan, Carlos D.  Castillo and Rama Chellappa}\\
    {\normalsize
    Center for Automation Research, UMIACS, University of Maryland, College Park, MD 20742\\}
    {\small \{rranjan1,swamiviv,carlos,rama\}@umiacs.umd.edu\\}}
}

\begin{document}

\maketitle

\begin{abstract}

We present a multi-purpose algorithm for simultaneous face detection, face alignment, pose estimation, gender recognition, smile detection, age estimation and face recognition using a single deep convolutional neural network (CNN). The proposed method employs a multi-task learning framework that regularizes the shared parameters of CNN and builds a synergy among different domains and tasks. Extensive experiments show that the network has a better understanding of face and achieves state-of-the-art result for most of these tasks.

\end{abstract}

\section{INTRODUCTION}
Face analysis is a challenging and actively researched problem with applications to face recognition, emotion analysis, biometrics security, etc. Though the performance of few challenging face analysis tasks such as unconstrained face detection and face verification have greatly improved when CNN-based methods are used, other tasks such as face alignment, head-pose estimation, gender and smile recognition are still challenging due to lack of large publicly available training data. Furthermore, all these tasks have been approached as separate problems, which makes their integration into end-to-end systems inefficient. For example, a typical face recognition system needs to detect and align a face from the given image before determining the identity. This results in error accumulation across different modules. Even though the above mentioned tasks are correlated, existing methods do not leverage the synergy among them. It has been shown recently that jointly learning correlated tasks can boost the performance of individual tasks~\cite{AFW_dataset_CVPR2012,DBLP:journals/corr/RanjanPC16,JointCascade_LI_ECCV2014}.

\begin{figure}[t]
      \centering
      \includegraphics[width=1\linewidth]{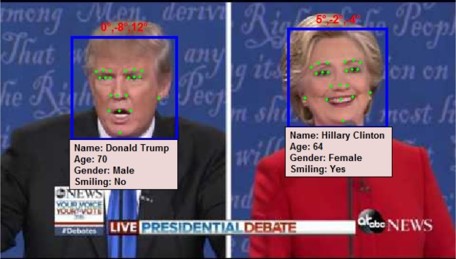}
      \caption{The proposed method can simultaneously detect faces, predict their landmarks locations, pose angles, smile expression, gender, age as well as the identity from any unconstrained face image.}
      \label{fig:front_page}
\end{figure}

In this paper, we present a novel CNN model that simultaneously solves the tasks of face detection, landmark localization, pose estimation, gender recognition, smile detection, age estimation and face verification and recognition (see Fig.~\ref{fig:front_page}). We choose this set of tasks since they span a wide range of applications. We train a CNN jointly in a multi-task learning (MTL) framework (Caruana~\cite{caruana1998multitask}), such that parameters from lower layers of CNN are shared among all the tasks. In this way, the lower layers learn general representation common to all the tasks, whereas upper layers are more specific to the given task, which reduces over-fitting in the shared layers. Thus, our model is able to learn robust features for distinct tasks. Employing multiple tasks enables the network to learn the correlations between data from different distributions in an effective way. This approach saves both time and memory in an end-to-end system, since it can simultaneously solve the tasks and requires the storage of a single CNN model instead of separate CNN for each task. To the best of our knowledge, this is the first work which simultaneously solves a diverse set of face analysis tasks using a single CNN in an end-to-end manner.


We initialize our network with the CNN model trained for face recognition task by Sankaranarayanan~et~al.~\cite{DBLP:journals/corr/Sankaranarayanan16a}. We argue that a network pre-trained on face recognition task possesses fine-grained information of a face which can be used to train other face-related tasks efficiently.  Task-specific sub-networks are branched out from different layers of this network depending on whether they rely on local or global information of the face. The complete network, when trained end-to-end, significantly improves the face recognition performance as well as other face analysis tasks. This paper makes the following contributions.
 
\begin{enumerate}
\item We propose a novel CNN architecture that simultaneously performs face detection, landmarks localization, pose estimation, gender recognition, smile detection, age estimation and face identification and verification.
\item We design a MTL framework for training, which regularizes the parameters of the network.
\item We achieve state-of-the-art performances on challenging unconstrained datasets for most of these tasks.
\end{enumerate}

Ranjan~et~al.~\cite{DBLP:journals/corr/RanjanPC16} recently proposed HyperFace that simultaneously performs the tasks of face detection, landmarks localization, pose estimation and gender recognition. The approach in this paper is different from HyperFace in the following aspects. Firstly, we additionally solve for the tasks of smile detection, age estimation and face recognition. Secondly, our MTL framework utilizes domain-based regularization by training on multiple datasets whereas HyperFace trains only on AFLW~\cite{AFLW}. Finally, we initialize our network with weights from face recognition task~\cite{DBLP:journals/corr/Sankaranarayanan16a} which provides a more robust and domain specific initialization while HyperFace network is initialized using the weights from AlexNet~\cite{NIPS2012_4824}. 

This paper is organized as follows. Section~\ref{sec:related} reviews closely related works.  Section~\ref{sec:proposed} describes the proposed algorithm in detail. Section~\ref{sec:experiments} provides the results of our method on challenging datasets for all the tasks. Finally, Section~\ref{sec:conclusion} concludes the paper with a brief discussion and future works.

\section{Related Work}
\label{sec:related}

Multitask learning was first analyzed in detail by Caruana~\cite{caruana1998multitask}. Since then, several approaches have used MTL for solving many problems in Computer Vision. One of the earlier methods for jointly learning face detection, landmarks localization and pose estimation was proposed in~\cite{AFW_dataset_CVPR2012} which was later extended to~\cite{FaceDPL}. It used a mixture of trees model with shared pool of parts, where a part represents a landmark location. Recently, several methods have incorporated the MTL framework with deep CNNs to train face-related tasks. Levi~et~al.~\cite{LH:CVPRw15:age} proposed a CNN for simultaneous age and gender estimation. HyperFace~\cite{DBLP:journals/corr/RanjanPC16} trained a MTL network for face detection, landmarks localization, pose and gender estimation by fusing the intermediate layers of CNN for improved feature extraction. Ehrlich~et~al.~\cite{ehrlich2016facial} proposed a multi-task restricted Boltzmann machine to learn facial attributes, while Zhang~et~al.~\cite{TCDCN} improved landmarks localization by training it jointly with head-pose estimation and facial attribute inference. Although these methods perform MTL on small set of tasks, they do not allow training a large set of correlated tasks as proposed in this paper.

Significant research has been undertaken for improving individual face analysis tasks. Recent methods for face detection based on deep CNNs such as DP2MFD~\cite{FD_BTAS2015}, Faceness~\cite{faceness_ICCV2015}, Hyperface~\cite{DBLP:journals/corr/RanjanPC16}, Faster-RCNN~\cite{jiang2016face}, etc., have significantly outperformed traditional approaches like TSM~\cite{AFW_dataset_CVPR2012} and NDPFace~\cite{NPDFace_PAMI2015}. 

Only a handful of methods have used deep CNNs for face alignment tasks~\cite{TCDCN,DBLP:journals/corr/KumarRPC16,zhu2015face,DBLP:journals/corr/RanjanPC16}, due to lack of sufficient training data.  Existing methods for landmark localization have focused mainly on near-frontal faces~\cite{DBLP:journals/ijcv/CaoWWS14,ren2014face,ERT} where all the essential keypoints are visible. Recent methods such as PIFA~\cite{Jourabloo_2015_ICCV}, 3DDFA~\cite{zhu2015face}, HyperFace~\cite{DBLP:journals/corr/RanjanPC16} and CCL~\cite{zhuunconstrained} have explored face alignment over varying pose angles. 

The task of pose estimation is to infer orientation of a person's head relative to the camera. Not much research has been carried out to solve this task for unconstrained images other than TSM~\cite{AFW_dataset_CVPR2012}, FaceDPL~\cite{FaceDPL} and HyperFace~\cite{DBLP:journals/corr/RanjanPC16}. 

The tasks of gender and smile classification from unconstrained images have been considered as a part of facial attribute inference. Recently, Liu~et~al.~\cite{CelebA} released CelebA dataset containing about $200,000$ near-frontal images with $40$ attributes including gender and smile, which accelerated the research in this field~\cite{wang2016walk,DBLP:journals/corr/RanjanPC16,PANDA,ehrlich2016facial}. Faces of the world~\cite{escalera2016chalearn} challenge dataset further advanced the research on these tasks for faces with varying scale, illumination and pose~\cite{li2016deepbe,uricarstructured,Zhang_2016_CVPR_Workshops}.

Age Estimation is the task of finding the real or apparent age of a person based on their face image. Few methods have already surpassed human error for the apparent age estimation challenge~\cite{escalera2015chalearn} using deep CNNs~\cite{Rothe-ICCVW-2015,chencascaded}. 
 
Face Verification is the task of predicting whether a pair of faces belong to the same person. Recent methods such as DeepFace~\cite{taigman2014deepface}, Facenet~\cite{schroff2015facenet}, VGG-Face~\cite{parkhi2015deep}  have significantly improved the verification accuracy on the LFW~\cite{LFWTech} dataset by training deep CNN models on millions of annotated data. However, it is still a challenging problem for unconstrained faces with large variations in viewpoint and illumination (IJB-A~\cite{klare2015pushing} dataset). We address this issue by regularizing the CNN parameters using the MTL framework, with only half-a-million samples (CASIA~\cite{DBLP:journals/corr/YiLLL14a}) for training.

\section{Proposed Method}
\label{sec:proposed}
We propose a multi-purpose CNN which can simultaneously detect faces, extract key-points and pose angles, determine smile expression, age and gender from any unconstrained image of a face. Additionally, it assigns an identity descriptor to each face which can be used for face recognition and verification. The proposed algorithm is trained in a MTL framework which builds a synergy among different face related tasks improving the performance for each of them. In this section we discuss the advantages of MTL in the context of face analysis and provide the details of the network design, training and testing procedures.

\subsection{Multi-task Learning}

\begin{figure}[htp!]
      \centering
      \includegraphics[width=6.0cm, height=3.0cm]{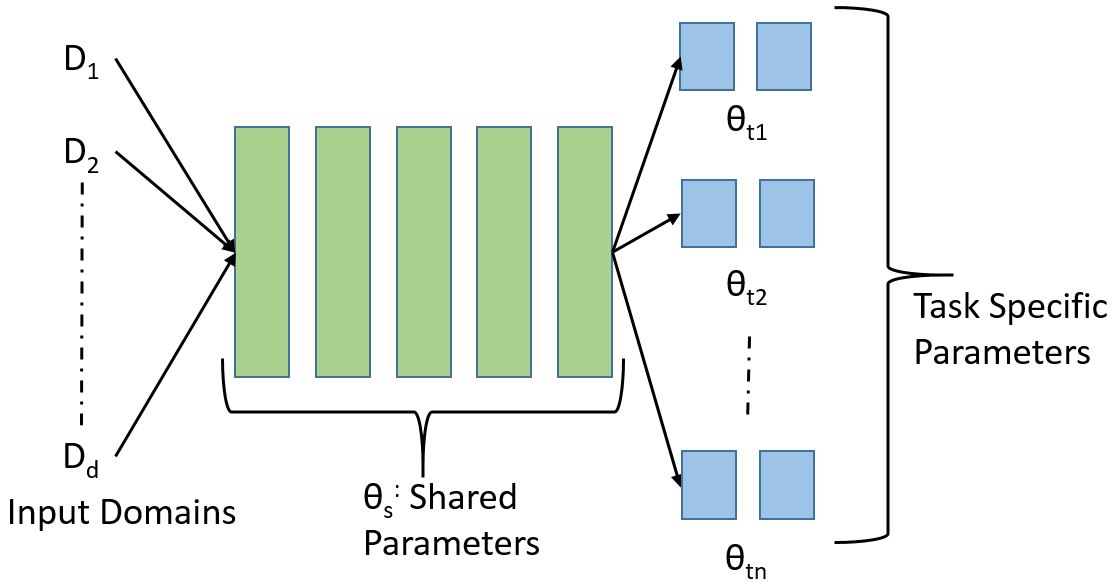}
      \caption{A general multitask learning framework for deep CNN architecture. The lower layers are shared among all the tasks and input domains.}
      \label{fig:multitask}
\end{figure}

Typically, a face analysis task requires a cropped face region as the input. The deep CNN processes the face to obtain a representation and extract meaningful information related to the task. According to~\cite{DBLP:journals/corr/ZeilerF13}, lower layers of CNN learn features common to a general set of face analysis tasks whereas upper layers are more specific to individual tasks. Therefore, we share the parameters of lower layers of CNN among different tasks to produce a generic face representation which is subsequently processed by the task-specific layers to generate the required outputs (Fig.~\ref{fig:multitask}). Goodfellow~et~al.~\cite{Goodfellow-et-al-2016-Book} interprets MTL as a regularization methodology for deep CNNs. The MTL approach used in our framework can be explained by following two types of regularization.

\subsubsection{Task-based Regularization}
Let the cost function for a given task~$t_{i}$ with shared parameters $\theta_{s}$ and task-specific parameters $\theta_{t_{i}}$ be $J_{i}(\theta_{s},\theta_{t_{i}};D)$, where $D$ is the input data. For isolated learning, the optimum network parameters $(\theta^{*}_{s},\theta^{*}_{t_{i}})$ can be computed using~(\ref{eq:opt_single_task})
\begin{equation}
\label{eq:opt_single_task}
(\theta^{*}_{s},\theta^{*}_{t_{i}}) = \argmin\limits_{(\theta_{s},\theta_{t_{i}})} J_{i}(\theta_{s},\theta_{t_{i}};D)
\end{equation}
For MTL, the optimal parameters for the task~$t_{i}$ can be obtained by minimizing the weighted sum of loss functions for each task, as shown in~(\ref{eq:opt_multi_task}). The loss weight for task~$t_{i}$ is denoted by $\alpha_{i}$. 
\begin{equation}
\label{eq:opt_multi_task}
(\theta^{*}_{s},\theta^{*}_{t_{i}}) = \argmin\limits_{(\theta_{s},\theta_{t_{i}})} \alpha_{i} J_{i}(\theta_{s},\theta_{t_{i}};D) + \sum\limits_{j \neq i}^n \alpha_{j} J_{j}(\theta_{s},\theta_{t_{j}};D)
\end{equation}
Since other tasks contribute only to the learning of shared parameters, they can be interpreted as a regularizer~$R_{i}$ on $\theta_{s}$ with respect to the given task~$t_{i}$, as shown in~(\ref{eq:opt_reg}). Thus, MTL shrinks the solution space of $\theta_{s}$ such that the learned parameter vector is in consensus with all the tasks, thus reducing over-fitting and enabling the optimization procedure to find a more robust solution. 
\begin{equation}
\label{eq:opt_reg}
(\theta^{*}_{s},\theta^{*}_{t_{i}}) = \argmin\limits_{(\theta_{s},\theta_{t_{i}})} J_{i}(\theta_{s},\theta_{t_{i}};D) + \lambda R_{i}(\theta_{s};D)
\end{equation}
\subsubsection{Domain-based Regularization}
For face analysis tasks, we do not have a large dataset with annotations for face bounding box, fiducial points, pose, gender, age, smile and identity information available. Hence, we adopt the approach of training multiple CNNs with respect to task-related datasets $D_{i}$, and share the parameters among them. In this way, the shared parameter $\theta_{s}$ adapts to the complete set of domains $(D_{1}, D_{2}, ... D_{d})$ instead of fitting to a task-specific domain. Additionally, the total number of training samples increases to roughly one-million, which is advantageous for training deep CNNs. Table~\ref{tbl:table_domain} lists the different datasets used for training our all-in-one CNN, along with their respective tasks and sample sizes.

\begin{table}
\caption{Datasets used for training}
\label{tbl:table_domain}
\begin{center}
\begin{tabular}{|c||c||c|}
\hline
Dataset & Face Analysis Tasks & \# training samples\\
\hline
CASIA~\cite{DBLP:journals/corr/YiLLL14a} & Identification, Gender & 490,356\\
\hline
MORPH~\cite{RicanekJr.:2006:MLI:1126250.1126361} & Age, Gender & 55,608\\
\hline
IMDB+WIKI~\cite{Rothe-ICCVW-2015} & Age, Gender & 224,840\\
\hline
Adience~\cite{LH:CVPRw15:age} & Age & 19,370\\
\hline
CelebA~\cite{CelebA} & Smile, Gender & 182,637\\
\hline
AFLW~\cite{AFLW} & Detection, Pose, Fiducials & 20,342\\
\hline
Total & & \textbf{993,153}\\
\hline
\end{tabular}
\end{center}
\end{table}

\subsection{Network Architecture}

\begin{figure}[htp!]
      \centering
      \includegraphics[width=7.0cm, height=10.0cm]{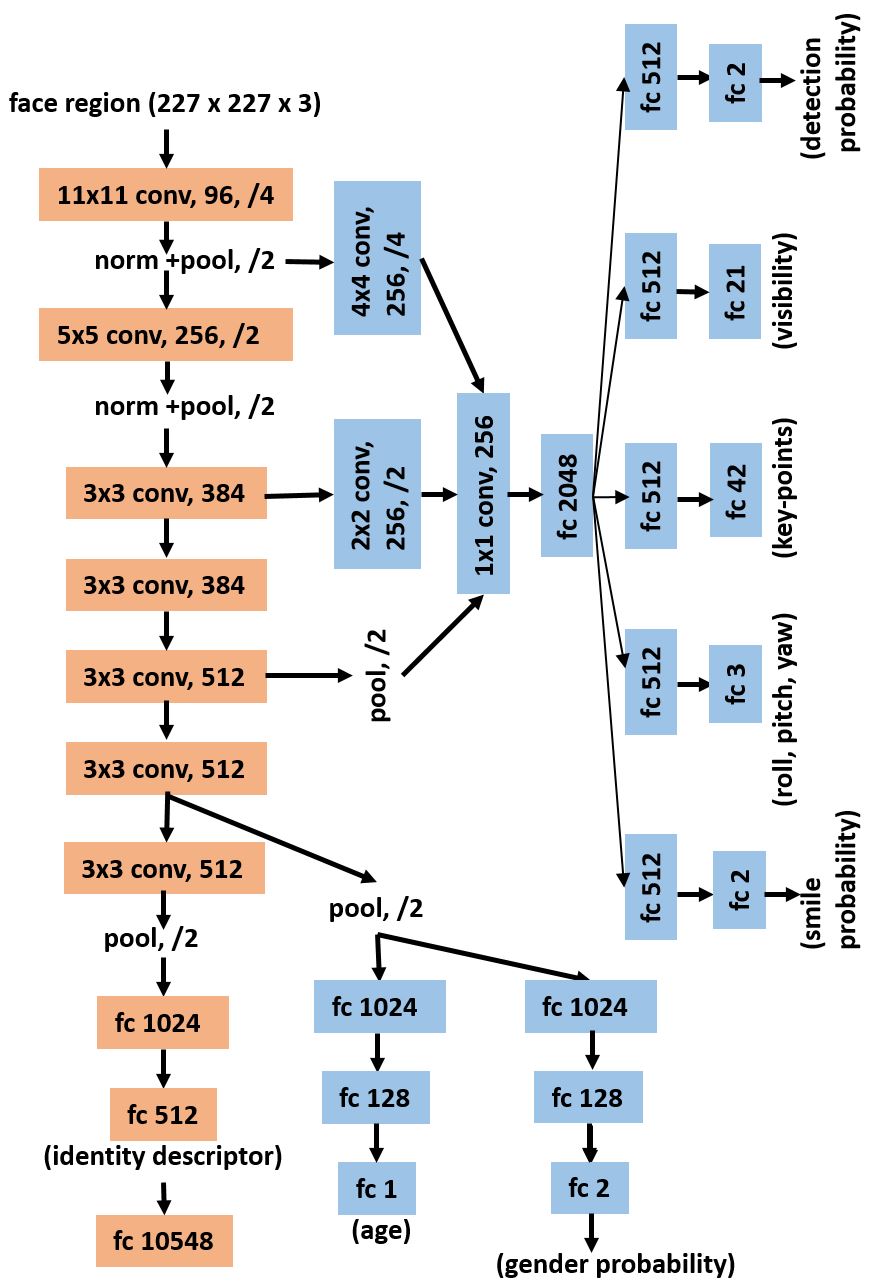}
      \caption{CNN Architecture for the proposed method. Each layer is represented by filter kernel size, type of layer, number of feature maps and the filter stride. Orange represents the pre-trained network from Sankaranarayanan~et~al.~\cite{DBLP:journals/corr/Sankaranarayanan16a}, while blue represents added layers for MTL.}
      \label{fig:architecture}
\end{figure}

The all-in-one CNN architecture is shown in Fig.~\ref{fig:architecture}. We start with the pre-trained face identification network from Sankaranarayanan~et~al.~\cite{DBLP:journals/corr/Sankaranarayanan16a}. The network consists of seven convolutional layers followed by three fully connected layers. We use it as a backbone network for training the face identification task and sharing the parameters from its first six convolution layers with other face-related tasks. Parametric Rectifier Linear units (PReLUs) are used as the activation function. We argue that a CNN pre-trained on face identification task provides a better initialization for a generic face analysis task, since the filters retain discriminative face information.

We divide the tasks into two groups: 1) subject-independent tasks which include face detection, keypoints localization and visibility,  pose estimation and smile prediction, and 2) subject-dependent tasks which include age estimation, gender prediction and face recognition. Similar to HyperFace~\cite{DBLP:journals/corr/RanjanPC16} we fuse the first, third and fifth convolutional layers for training the subject-independent tasks, as they rely more on local information available from the lower layers of the network. We attach two convolution layers and a pooling layer respectively to these layers, to obtain a consistent feature map size of $6 \times 6$. A dimensionality reduction layer is added to reduce the number of feature maps to $256$. It is followed by a fully connected layer of dimension $2048$, which forms a generic representation for subject-independent tasks. At this point, the specific tasks are branched into fully connected layers of dimension $512$ each, which are followed by the output layers respectively.

The subject-dependent tasks of age estimation and gender classification are branched out from the sixth convolutional layer of the backbone network after performing the max pooling operation. The global features thus obtained are fed to a $3$-layered fully connected network for each of these tasks. We keep the seventh convolutional layer unshared to adapt it specifically to the face recognition task. Task-specific loss functions are used to train the complete network end-to-end.

\subsection{Training}

The training CNN model contains five sub-networks with parameters shared among them. The tasks of face detection, key-points localization and visibility, and pose estimation are trained in a single sub-network, since all of them use a common dataset (AFLW~\cite{AFLW}) for training. The remaining tasks of smile detection, gender recognition, age estimation and face recognition are trained using separate sub-networks. At test time, these sub-networks are fused together into a single all-in-one CNN~(Fig.~\ref{fig:architecture}). All tasks are trained end-to-end simultaneously using Caffe~\cite{jia2014caffe}. Here, we discuss the loss functions and training dataset for each of them.


\subsubsection{Face Detection, Key-points Localization and Pose Estimation} These tasks are trained in a similar manner as HyperFace~\cite{DBLP:journals/corr/RanjanPC16}, using AFLW~\cite{AFLW} dataset. We randomly select $1000$ images from the dataset for testing, and use the remaining images for training. We use the Selective Search~\cite{vandeSande:2011:SSS:2355573.2356474} algorithm to generate region proposals for faces from an image. Regions with Intersection-Over-Union (IOU) overlap of more than $0.5$ with the ground truth bounding-box are considered positive examples whereas regions with IOU\textless$0.35$ are chosen as negative examples for training the detection task using a softmax loss function. Facial landmarks, key-points visibility and pose estimation tasks are treated as regression problems and trained with the Euclidean loss. Only regions with IOU\textgreater$0.35$ contribute to back-propagation during their training.

\subsubsection{Gender Recognition}
It is a binary classification problem similar to face detection. The datasets used for training gender are listed in Table~\ref{tbl:table_domain}. The training images are first aligned using facial key-points which are either provided by the dataset or computed using HyperFace~\cite{DBLP:journals/corr/RanjanPC16}. A cross-entropy loss $L_{G}$ is used for training as shown in~(\ref{eq:loss_gender}) 

\begin{equation}
\label{eq:loss_gender}
L_{G} = -(1-g) \cdot \log(1-p_{g})-g \cdot \log(p_{g}),
\end{equation}

where $g = 0$ for male and $1$ for female. $p_{g}$ is the predicted probability that the input face is a female.

\subsubsection{Smile Detection} The smile attribute is trained  to make the network robust to expression variations for face recognition. We use CelebA~\cite{CelebA} dataset for training. Similar to the gender classification task, the the images are aligned  before passing them through the network. The loss function $L_{S}$ is given by (\ref{eq:loss_smile})

\begin{equation}
\label{eq:loss_smile}
L_{S} = -(1-s) \cdot \log(1-p_{s})-s \cdot \log(p_{s}),
\end{equation}

where $s=1$ for a smiling face and $0$ otherwise. $p_{s}$ is the predicted probability that the input face is a smiling.

\subsubsection{Age Estimation}
\begin{figure}[htp!]
      \centering
      \includegraphics[width=5.0cm, height=3.0cm]{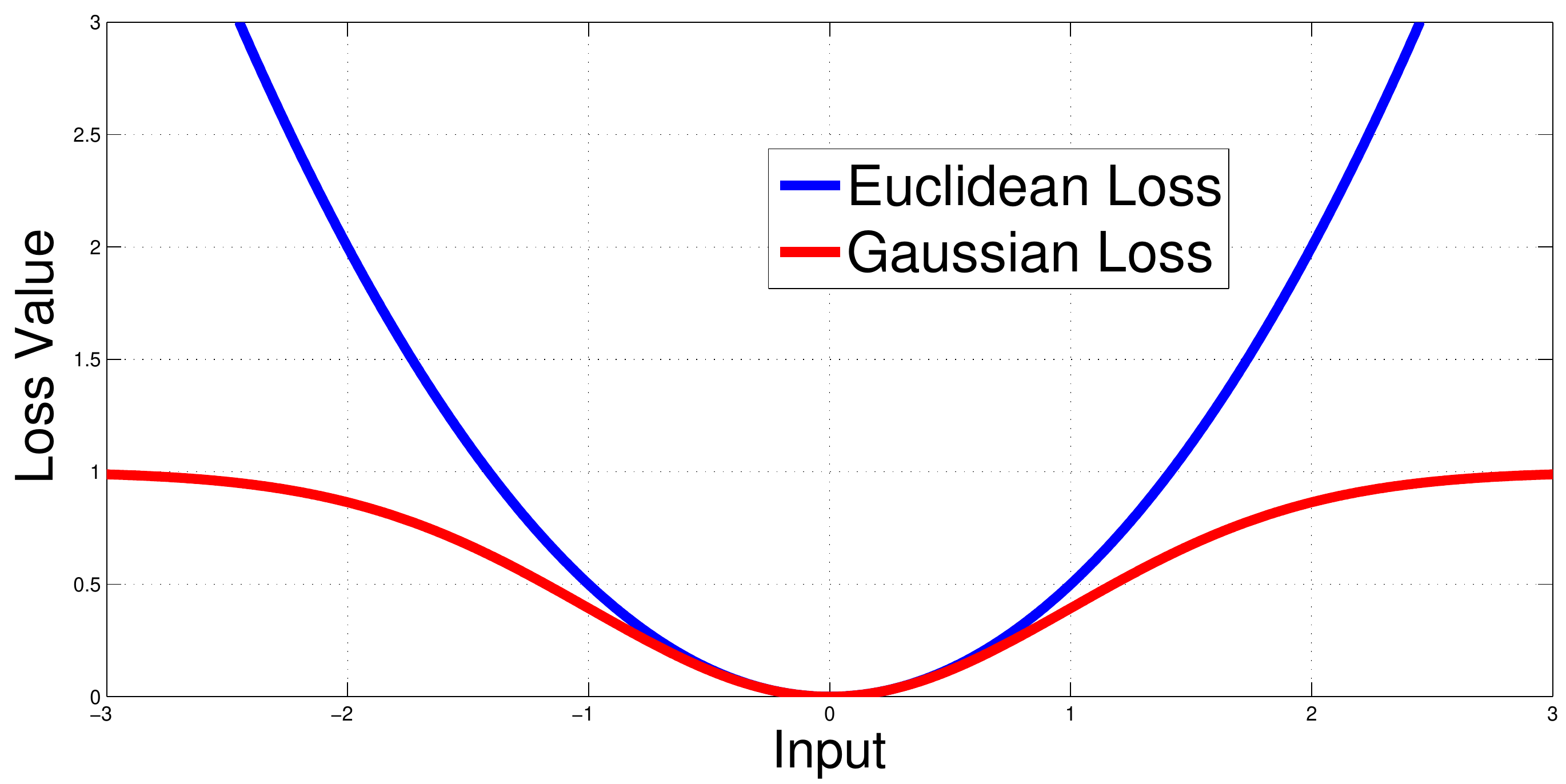}
      \caption{Euclidean and Gaussian loss functions.}
      \label{fig:euc_gauss}
\end{figure}

We formulate the age estimation task as a regression problem in which the network learns to predict the age from a face image. We use IMDB+WIKI~\cite{Rothe-ICCVW-2015}, Adience~\cite{LH:CVPRw15:age} and MORPH~\cite{RicanekJr.:2006:MLI:1126250.1126361} datasets for training. It has been shown by Ranjan~et.~al.~\cite{Ranjan:2015:UAE:2919341.2921043} that Gaussian loss works better than Euclidean loss for apparent age estimation when the standard deviation of age is given. However, the gradient of Gaussian loss is close to zero when the predicted age is far from the true age (Fig.~\ref{fig:euc_gauss}), which slows the training process. Hence, we use a linear combination of these two loss functions weighted by $\lambda$ as shown in~(\ref{eq:loss_age}) 

\begin{equation}
\label{eq:loss_age}
L_{A} = (1-\lambda) \frac{1}{2}(y-a)^{2} + \lambda \left(1-exp(-\frac{(y-a)^{2}}{2\sigma^{2}})\right),
\end{equation}

where $L_{A}$ is the age loss, $y$ is the predicted age, $a$ is the ground-truth age and $\sigma$ is the standard deviation of the annotated age value. $\lambda$ is initialized with $0$ at the start of the training, and increased to $1$ subsequently. In our implementation, we keep $\lambda = 0$ initially and switch it to $1$ after $20k$ iterations. $\sigma$ is fixed at $3$ if not provided by the training set.

\subsubsection{Face Recognition}
We use $10,548$ subjects from CASIA~\cite{DBLP:journals/corr/YiLLL14a} dataset to train the face identification task. The images are aligned using HyperFace~\cite{DBLP:journals/corr/RanjanPC16} before passing them through the network. We deploy a multi-class cross-entropy loss function $L_{R}$ for training as shown in (\ref{eq:loss_recog})

\begin{equation}
\label{eq:loss_recog}
L_{R} = \sum \limits_{c=0}^{10547} -y_{c} \cdot \log(p_{c}),
\end{equation}

where $y_{c} = 1$ if the sample belongs to class $c$, otherwise $0$. The predicted probability that a sample belongs to class $c$ is given by $p_{c}$.

The final overall loss $L$ is the weighted sum of individual loss functions, given in~(\ref{eq:loss_all})

\begin{equation}
\label{eq:loss_all}
L = \sum_{t=1}^{t=8}\lambda_{t}L_{t},
\end{equation}

where $L_{t}$ is the loss and $\lambda_{t}$ is the loss-weight corresponding to task $t$. The loss-weights are chosen empirically. We assign a higher weight to regression tasks as they tend to have lower loss magnitude than classification tasks.

\subsection{Testing}

\begin{figure}[htp!]
      \centering
      \includegraphics[width=1\linewidth]{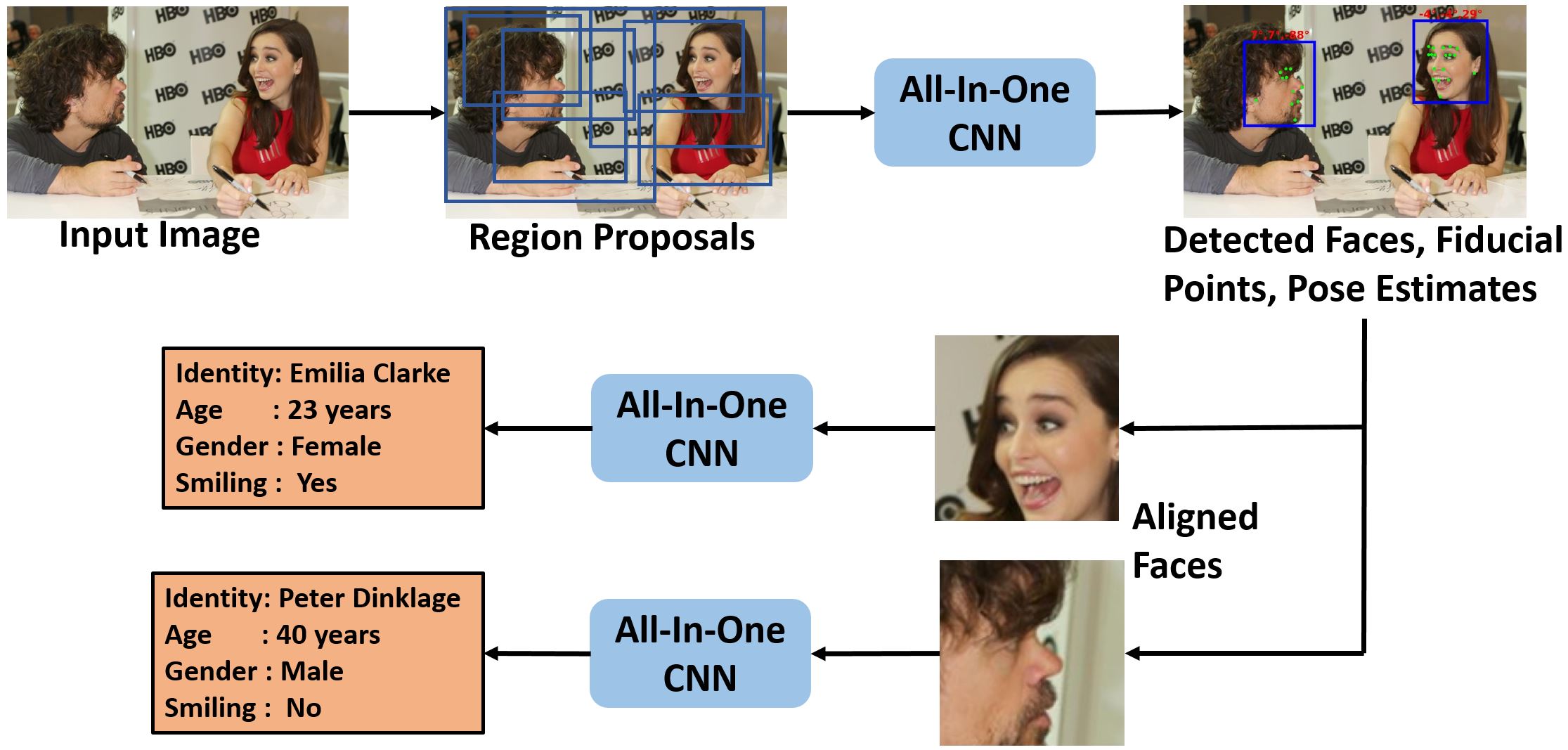}
      \caption{The end-to-end pipeline for the proposed method during test time.}
      \label{fig:pipeline}
\end{figure}

We deploy a two-stage process during test time as shown in Fig.~\ref{fig:pipeline}. In the first stage, we use the Selective Search~\cite{vandeSande:2011:SSS:2355573.2356474} to generate region proposals from a test image, which are passed through our all-in-one network to 
obtain the detection scores, pose estimates, fiducial points and their visibility. We use Iterative Region Proposals and Landmarks-based NMS~\cite{DBLP:journals/corr/RanjanPC16} to filter out non-faces and improve fiducials and pose estimates. 

For the second stage, we use the obtained fiducial points to align each detected face to a canonical view using similarity transform. 
The aligned faces, along with their flipped versions are passed again through the network to get the smile, gender, age and identity information. We use the $512$-dimensional feature from the penultimate fully connected layer of the identification network as the identity descriptor.

\section{Experiments}
\label{sec:experiments}
The proposed method is evaluated for all the tasks on which it was trained except the key-points visibility, due to the lack of a proper evaluation protocol. We select HyperFace~\cite{DBLP:journals/corr/RanjanPC16} as a comparison baseline for the tasks of face detection, pose estimation, landmarks localization and gender recognition. For face recognition task, the method from Sankaranarayanan~et.~al.~\cite{DBLP:journals/corr/Sankaranarayanan16a}, which is used as the initialization, is used as the baseline.


\begin{figure*}[htp!]
 \centering
\includegraphics[width=6.0cm, height=3.4cm]{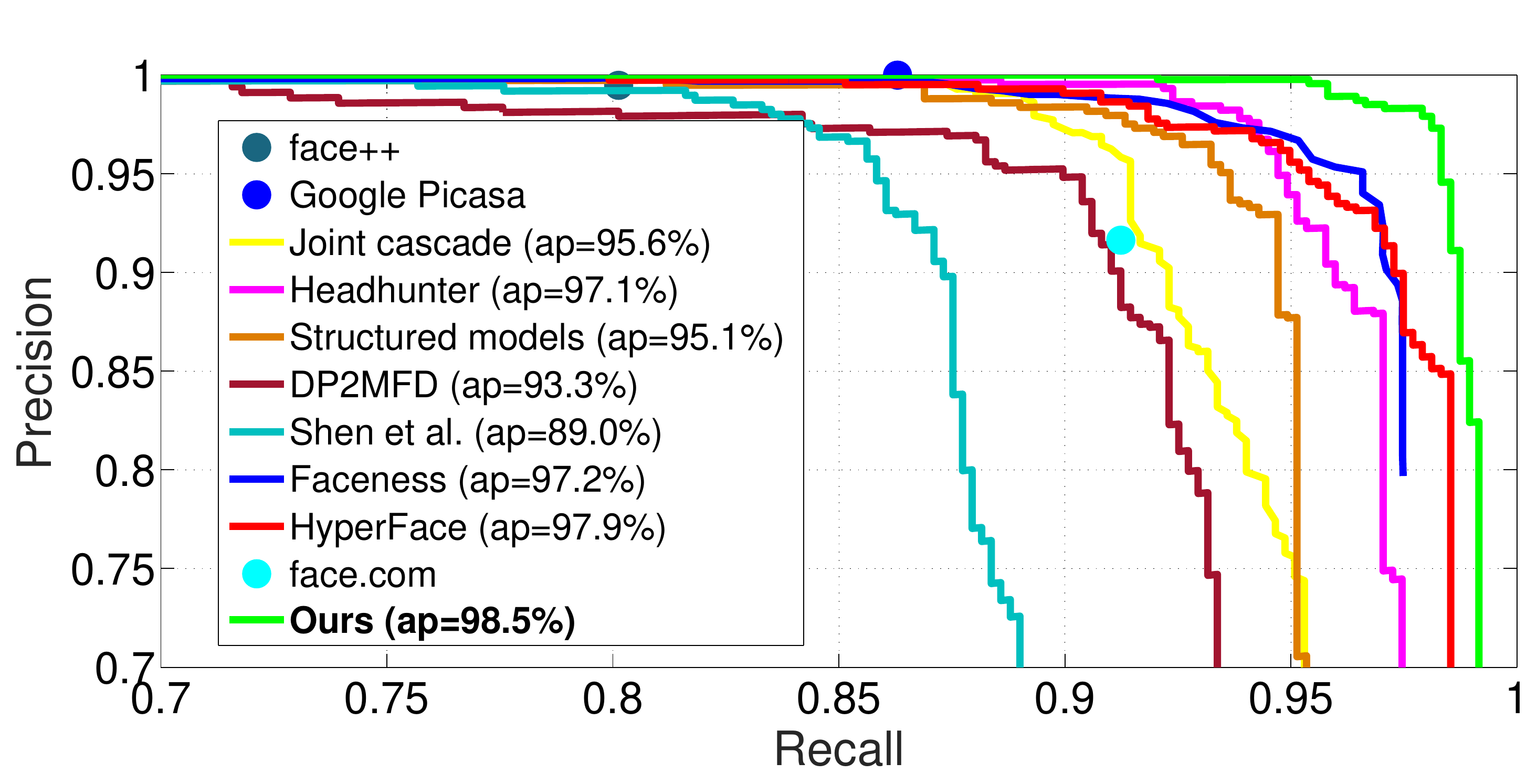}\hskip20pt\includegraphics[width=5.0cm, height=3.2cm]{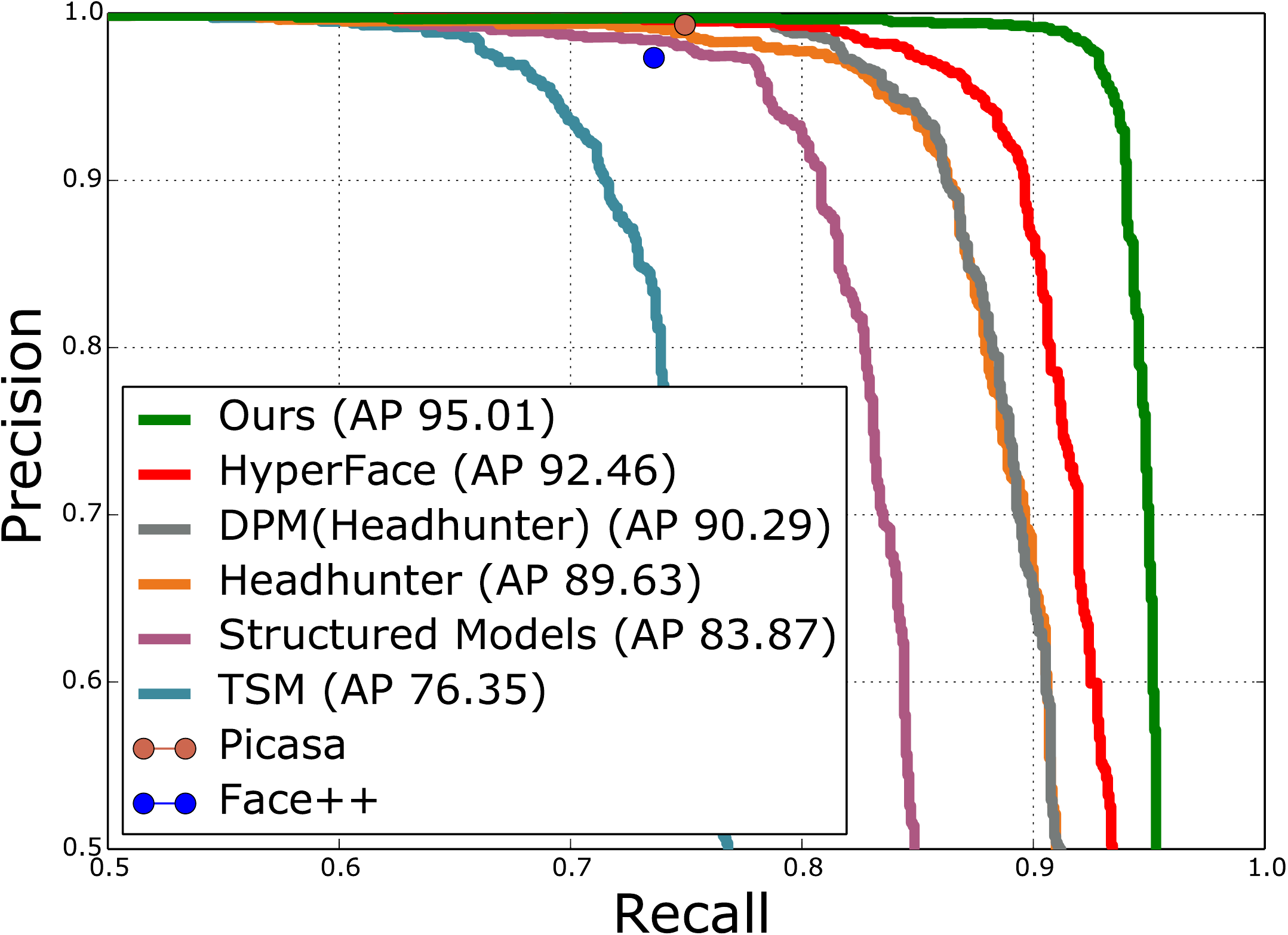}
\hskip20pt\includegraphics[width=5.0cm, height=3.2cm]{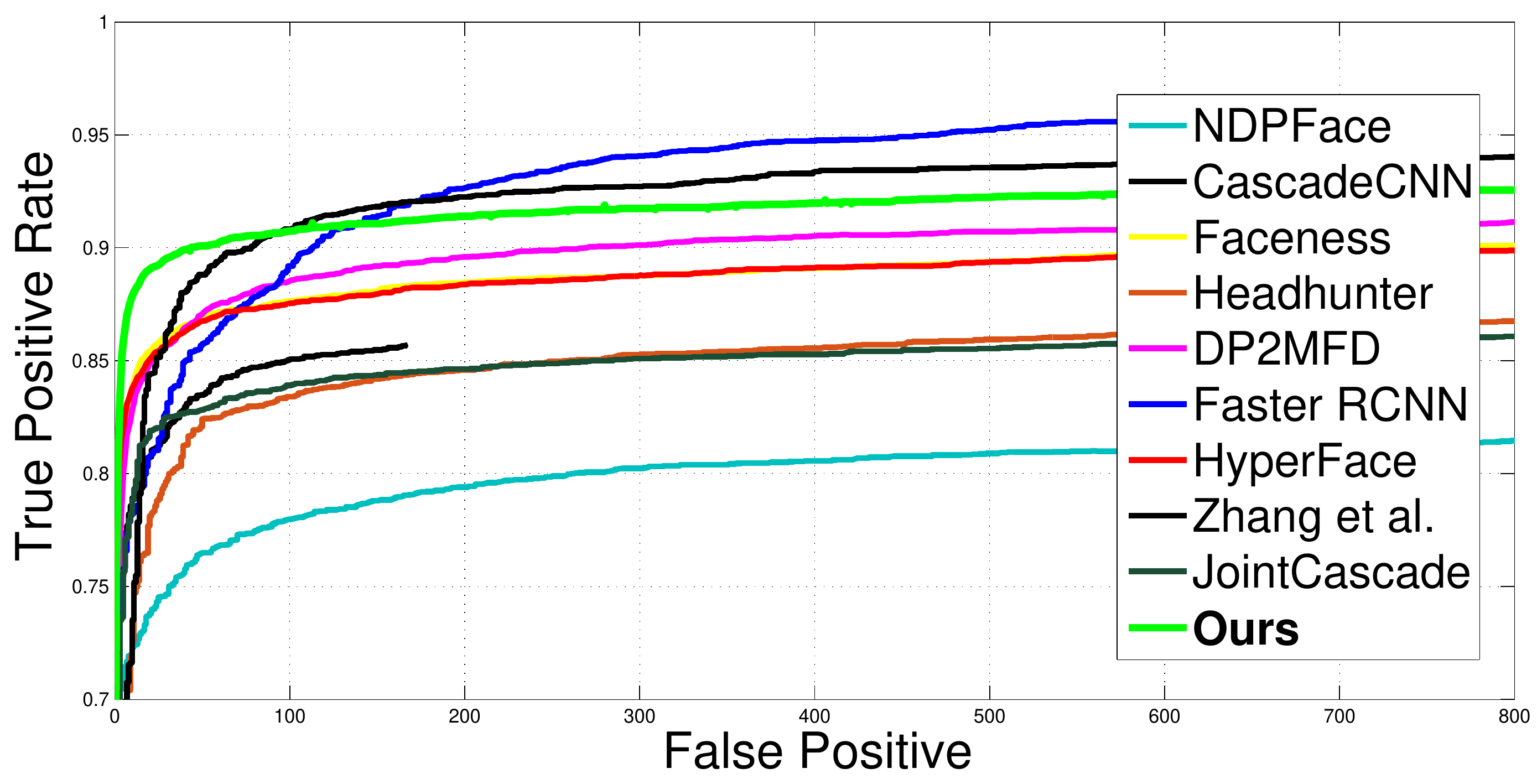}\\
\hskip25pt(a)\hskip165pt(b)\hskip155pt(c)
\caption{Performance evaluation on (a) the AFW dataset, (b) the PASCAL faces dataset and (c) FDDB dataset. The numbers in the legend are the mean average precision for the corresponding datasets.}
\label{fig:detections}
\end{figure*}

\subsection{Face Detection}
We evaluate the face detection task on Annotated Face in-the-Wild (AFW)~\cite{AFW_dataset_CVPR2012}, PASCAL faces~\cite{PASCAL_faces} and Face Detection Dataset and Benchmark (FDDB)~\cite{fddbTech} datasets. All these datasets contain faces with wide variations in appearance, scale, viewpoint and illumination. To evaluate on AFW and PASCAL datasets, we finetune the face detection branch of the network on FDDB. To evaluate on the FDDB dataset, we finetune according to the 10-fold cross validation experiments~\cite{fddbTech}.


The precision-recall curves for AFW and PASCAL dataset, and Receiver Operating Characteristic (ROC) curve for FDDB dataset are shown in Fig.~\ref{fig:detections}. It can be seen from the figures that our method achieves state-of-the-art performance on AFW and PASCAL dataset with mean average precision (mAP) of $98.5\%$ and $95.01\%$ respectively. On FDDB dataset, our method performs better than most of the reported algorithms. It gets lower recall than Faster-RCNN~\cite{jiang2016face} and Zhang~et~al.~\cite{zhang2016joint}, since small faces of FDDB fail to get captured in any of the region proposals. Other recently published methods compared in our detection evaluations include DP2MFD~\cite{FD_BTAS2015}, Faceness~\cite{faceness_ICCV2015}, Headhunter~\cite{HeadHunter_Mathias_ECCV2014}, Joint Cascade~\cite{JointCascade_LI_ECCV2014}, Structured Models~\cite{PASCAL_faces}, Cascade CNN~\cite{CascadeCNN_CVPR2015}, NDPFace~\cite{NPDFace_PAMI2015}, TSM~\cite{AFW_dataset_CVPR2012}, as well as three commercial systems Face++, Picasa and Face.com.

\subsection{Landmarks Localization}
We evaluate our performance on AFW~\cite{AFW_dataset_CVPR2012} and AFLW~\cite{AFLW} datasets as they contain large variations in viewpoints of faces. The landmarks location is computed as the mean of the predicted landmarks corresponding to region proposals having IOU\textgreater 0.5 with the test face. For AFLW~\cite{AFLW} evaluation, we follow the protocol given in~\cite{zhu2015face}. We randomly create a subset of $450$ samples from our test set such that the absolute yaw angles within [$0^{\circ}, 30^{\circ}$], [$30^{\circ}, 60^{\circ}$] and [$60^{\circ}, 90^{\circ}$] are $1/3$ each. Table~\ref{tbl:aflw} compares the Normalized Mean Error (NME) for our method with recent face alignment method adapted to face profoling~\cite{zhu2015face}, for each of the yaw bins. Our method significantly outperforms the previous best HyperFace~\cite{DBLP:journals/corr/RanjanPC16}, reducing the error by more than $30\%$. A low standard deviation of $0.13$ suggests that landmarks prediction is consistent as pose angles vary.

\begin{table}[htp!]
\centering
\caption{The NME(\%) of face alignment results on AFLW test set.}
\label{tbl:aflw}
\begin{tabular}{|c|c|c|c|c|c|}
\hline
                   & \multicolumn{5}{c|}{AFLW Dataset (21 pts)}                                    \\ \hline
Method             & {[}0, 30{]}   & {[}30, 60{]}  & {[}60, 90{]}  & mean          & std           \\ \hline
RCPR~\cite{burgos2013robust}              & 5.43          & 6.58          & 11.53         & 7.85          & 3.24          \\ \hline
ESR~\cite{DBLP:journals/ijcv/CaoWWS14}                & 5.66          & 7.12          & 11.94         & 8.24          & 3.29          \\ \hline
SDM~\cite{XiongD13}               & 4.75          & 5.55          & 9.34          & 6.55          & 2.45          \\ \hline
3DDFA~\cite{zhu2015face}              & 5.00          & 5.06          & 6.74          & 5.60          & 0.99          \\ \hline
3DDFA+SDM          & 4.75          & 4.83          & 6.38          & 5.32          & 0.92          \\ \hline
HyperFace~\cite{DBLP:journals/corr/RanjanPC16} 		   & 3.93          & 4.14          & 4.71          & 4.26 			& 0.41          \\ \hline
\textbf{Ours} & \textbf{2.84} & \textbf{2.94} & \textbf{3.09}          & \textbf{2.96} 	& \textbf{0.13}          \\ \hline
\end{tabular}
\end{table}

For AFW~\cite{AFW_dataset_CVPR2012} evaluation, we follow the protocol described in~\cite{zhuunconstrained}. Fig.~\ref{fig:landmark_pose}(a) shows comparisons with recently published methods such as CCL~\cite{zhuunconstrained}, HyperFace~\cite{DBLP:journals/corr/RanjanPC16}, LBF~\cite{ren2014face}, SDM~\cite{XiongD13}, ERT~\cite{ERT} and  RCPR~\cite{burgos2013robust}. It is evident that the proposed algorithm performs better than existing methods on unconstrained and profile faces since it predicts landmarks with less than $5\%$ NME on more than $95.5\%$ of test faces. However, it lacks in pixel-accurate precise localization of key-points for easy faces, which can be inferred from the lower end of the curve. Most of these algorithms use cascade stage-wise regression to improve the localization, which is slower compared to a single forward pass of the network.

\subsection{Pose Estimation}

\begin{figure}[htp!]
 \centering
\includegraphics[width=4.0cm, height=2.7cm]{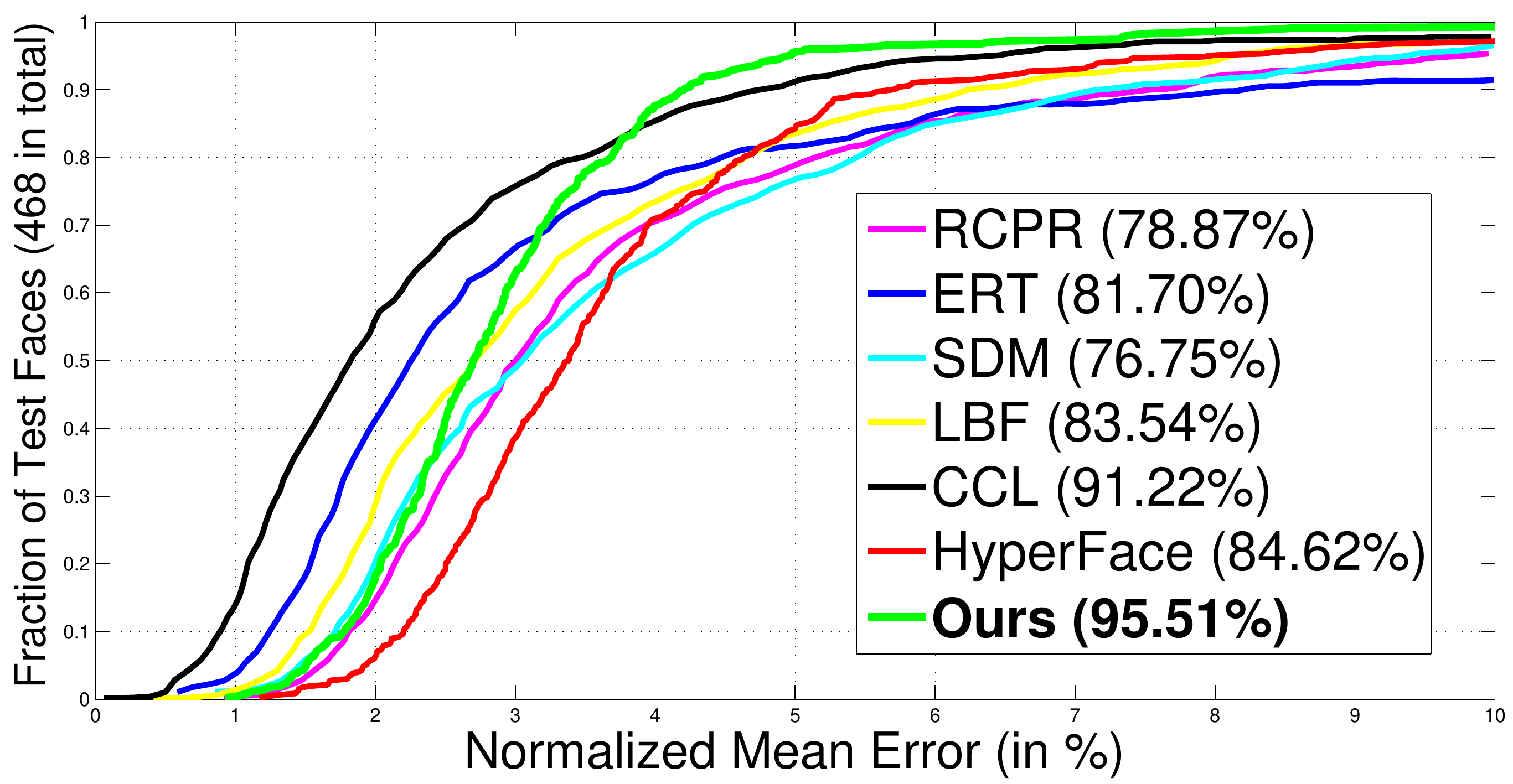}\hskip10pt\includegraphics[width=4.0cm, height=2.7cm]{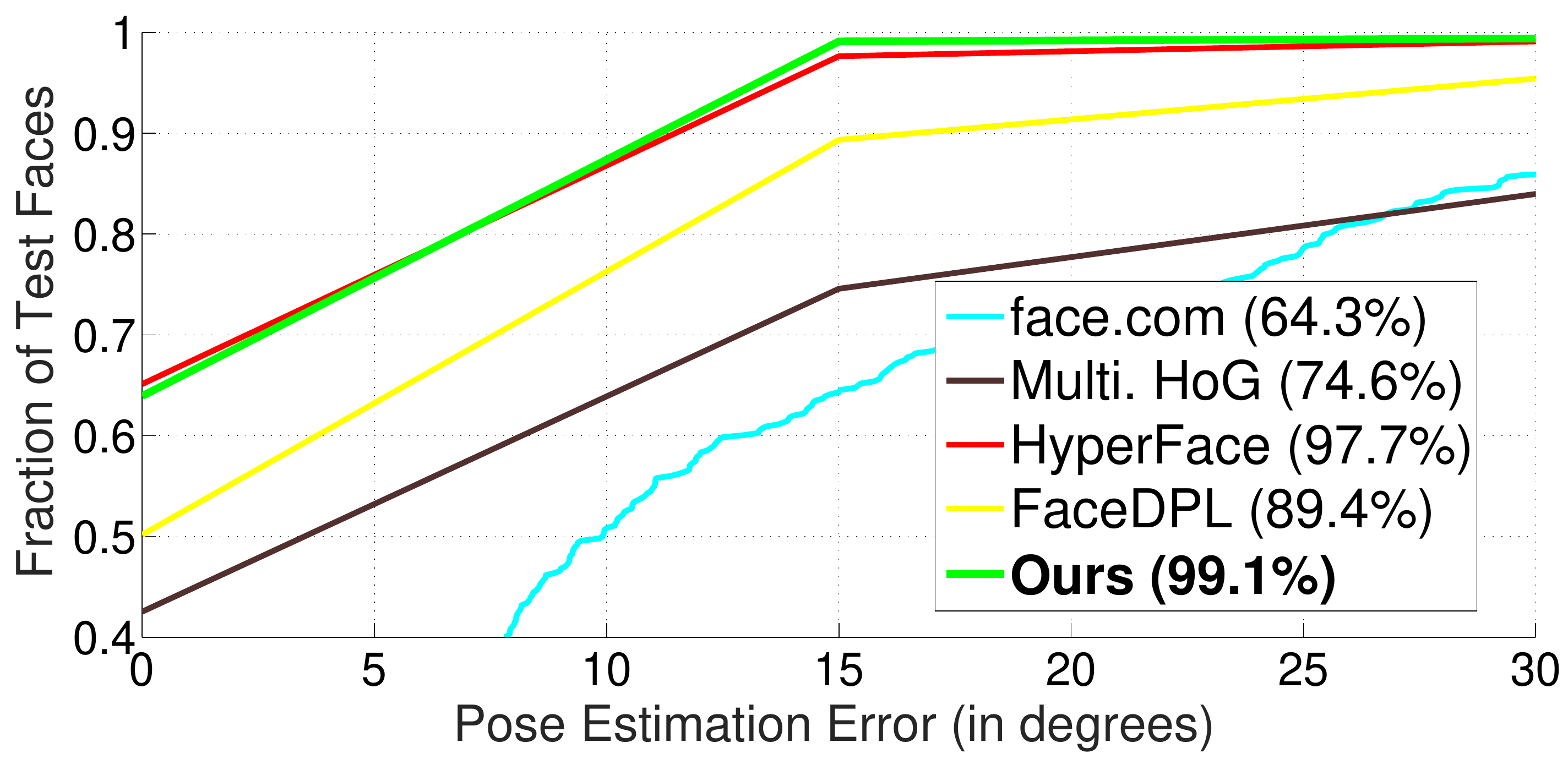}\\
(a)\hskip120pt(b)\\
\caption{Performance evaluation on AFW dataset for (a) landmarks localization task, (b) pose estimation task. The numbers in the legend are the percentage of test faces with (a) NME less than $5\%$, (b) absolute yaw error less than or equal to $15^{\circ}$}
\label{fig:landmark_pose}
\end{figure}

We evaluate our metod on AFW~\cite{AFW_dataset_CVPR2012} dataset for the pose estimation task. According to the protocol defined in~\cite{AFW_dataset_CVPR2012}, we compute the absolute error only for the yaw angles. Since, the ground-truth yaw angles are provided in multiples of $15^{\circ}$, we round-off our predicted yaw to the nearest $15^{\circ}$ for evaluation. Fig.~\ref{fig:landmark_pose}(b) shows the comparison of our method with HyperFace~\cite{DBLP:journals/corr/RanjanPC16}, Face DPL~\cite{FaceDPL}, Multiview HoG \cite{AFW_dataset_CVPR2012} and face.com. It is clear that the proposed algorithm performs better than competing methods and is able to predict the yaw in the range of $\pm 15^{\circ}$ for more than $99\%$ of the faces.

\subsection{Gender and Smile Recognition}

We evaluate the smile and gender recognition performance on Large-scale CelebFaces Attributes (CelebA)~\cite{CelebA} and ChaLearn Faces of the World~\cite{escalera2016chalearn} datasets. While CelebA~\cite{CelebA} has wide variety of subjects, they mostly contain frontal faces. Faces of the World~\cite{escalera2016chalearn} has wide variations in scale and viewpoints of faces.  We take the mean of the predicted scores obtained from region proposals having IOU\textgreater 0.5 with the given face, as our final score for smile and gender attributes. Table~\ref{tbl:gender_smile} compares the gender and smile accuracy with recently published methods. On CelebA~\cite{CelebA}, we outperform all the methods for gender accuracy. Our smile accuracy is lower only to Walk and Learn~\cite{wang2016walk} which uses other contextual information to improve the prediction.  The gender and smile branches of the network were finetuned on the training set of Faces of the World~\cite{escalera2016chalearn} before its evaluation. We achieve state-of-the-art performance for both gender and smile classification on their validation set. 

\begin{table}[htp!]
\caption{Accuray (\%) for Gender and Smile classification on CelebA~\cite{CelebA} (left) and Faces of the World~\cite{escalera2016chalearn} (right)}
\label{tbl:gender_smile}
\begin{minipage}{.25\linewidth}
\tabcolsep=0.08cm
\begin{tabular}{|c|c|c|}
\hline
Method & Gender&Smile\\
\hline\hline
PANDA-1~\cite{PANDA}&97&92\\
MT-RBM~\cite{ehrlich2016facial}& 90&88\\
LNets+ANet~\cite{CelebA} &98&92\\
HyperFace~\cite{DBLP:journals/corr/RanjanPC16}&97&-\\
Walk \& Learn~\cite{wang2016walk}&96&\bf{98}\\
\textbf{Ours}&\bf{99}&93\\
\hline
\end{tabular}
\end{minipage}%
\hskip60pt
\begin{minipage}{.25\linewidth}
\tabcolsep=0.08cm
\begin{tabular}{|c|c|c|}
\hline
Method & Gender&Smile\\
\hline\hline
MT-RBM~\cite{ehrlich2016facial}&71.7&80.8\\
CMP+ETH~\cite{uricarstructured}&89.15&79.03\\
DeepBE~\cite{li2016deepbe}&90.44&88.43\\
SIAT\_MMLAB~\cite{Zhang_2016_CVPR_Workshops}&91.66&89.34\\
\textbf{Ours}&\bf{93.12}&\bf{90.83}\\
\hline
\end{tabular}
\end{minipage}
\end{table}

\subsection{Age Estimation}
We use Chalearn LAP2015~\cite{escalera2015chalearn} apparent age estimation challenge dataset and FG-NET~\cite{han2013age} Aging Database for evaluating our age estimation task. We fine-tune the age-task branch of the network on the training set of the challenge dataset, and show the results on the validation set. The error is computed according to the protocol described in~\cite{escalera2015chalearn}.  For FG-Net~\cite{han2013age}, we follow the standard Leave-One-Out-Protocol (LOPO). Table~\ref{tbl:age} lists the evaluation error for both these datasets. We surpass human error of $0.34$ and perform comparable to state-of-the-art methods, obtaining an error of $0.293$ on Chalearn LAP2015~\cite{escalera2015chalearn} dataset. On FG-Net~\cite{han2013age}, we significantly outperform other methods, achieving an average error of $2$ years.

\begin{table}[htp!]
\centering
\caption{Age Estimation error on LAP2015(left) and FG-NET(right)}
\label{tbl:age}
\begin{minipage}{.25\linewidth}
\tabcolsep=0.15cm
\begin{tabular}{|c|c|}
\hline
Method & Error\\
\hline\hline
UMD~\cite{Ranjan:2015:UAE:2919341.2921043}&0.359\\
Human&0.34\\
CascadeAge~\cite{chencascaded}&0.297\\
CVL\_ETHZ&0.295\\
ICT-VIPL&\bf{0.292}\\
Ours&0.293\\
\hline
\end{tabular}
\end{minipage}%
\hskip50pt
\begin{minipage}{.25\linewidth}
\tabcolsep=0.15cm
\begin{tabular}{|c|c|}
\hline
Method & Error\\
\hline\hline
Han2013~\cite{han2013age}&4.6\\
Chao2013~\cite{chao2013facial}&4.38\\
Hong2013~\cite{hong2013new}&4.18\\
El Dib2010~\cite{el2010human}&3.17\\
CascadeAge~\cite{chencascaded}&3.49\\
\textbf{Ours}&\bf{2.00}\\
\hline
\end{tabular}
\end{minipage}
\end{table}

\subsection{Face Identification/Verification}
We evaluate the tasks of face recognition and verification on the IARPA Janus Benchmark-A (IJB-A)~\cite{klare2015pushing} dataset. The dataset contains $500$ subjects with a total of $25,813$ images including $5,399$ still images and $20,414$ video frames. It contains faces with extreme viewpoints, resolution and illumination which makes it more challenging than the commonly used LFW~\cite{LFWTech} dataset.

\begin{table*}
\caption{Face Identification and Verification Evaluation on IJB-A dataset}
\label{tbl:ijba}
\begin{center}
\tabcolsep=0.15cm
\begin{tabular}{|c||c||c||c||c||c||c||c|}
\hline
 & \multicolumn{3}{c|}{IJB-A Verification (TAR@FAR)} & \multicolumn{4}{c|}{IJB-A Identification}\\
\hline
Method & 0.001 & 0.01 & 0.1 & FPIR=0.01 & FPIR=0.1 & Rank=1 & Rank=10\\
\hline
GOTS~\cite{klare2015pushing} & 0.2(0.008) & 0.41(0.014) & 0.63(0.023) & 0.047(0.02) & 0.235(0.03) & 0.443(0.02) & -\\
VGG-Face~\cite{parkhi2015deep} & 0.604(0.06) & 0.805(0.03) & 0.937(0.01) & 0.46(0.07) & 0.67(0.03) & 0.913(0.01) & 0.981(0.005)\\
Chen~et~al.~\cite{chen2016unconstrained} & - & 0.838(0.042) & 0.967(0.009) & - & - & 0.903(0.012) & 0.977(0.007)\\
Masi
et al.~\cite{masi2016we} & 0.725 & 0.886 & - & - & - & 0.906 & 0.977\\
NAN~\cite{DBLP:journals/corr/YangRCWLH16}& 0.785(0.03) & 0.897(0.01) & 0.959(0.005) & - & - & - & -\\
Sankaranarayanan~et~al.~\cite{DBLP:journals/corr/Sankaranarayanan16a} w/o TPE & 0.766(0.02) & 0.871(0.01) & 0.952(0.005) & 0.67(0.05) & 0.82(0.013) & 0.925(0.01) & 0.978(0.005)\\
Sankaranarayanan~et~al.~\cite{DBLP:journals/corr/Sankaranarayanan16a} & 0.813(0.02) & 0.90(0.01) & 0.964(0.005) & 0.753(0.03) & 0.863(0.014) & 0.932(0.01) & 0.977(0.005)\\
Crosswhite~et~al.~\cite{crosswhite2016template} & - & \bf{0.939(0.013)} & - & 0.774(0.049) & 0.882(0.016) & 0.928(0.01) & 0.986(0.003)\\
\textbf{Ours} & 0.787(0.04) & 0.893(0.01) & 0.968(0.006) & 0.704(0.04) & 0.836(0.014) & 0.941(0.008) & 0.988(0.003)\\
\textbf{Ours + TPE~\cite{DBLP:journals/corr/Sankaranarayanan16a}} & \bf{0.823(0.02)} & 0.922(0.01) & \bf{0.976(0.004)} & \bf{0.792(0.02)} & \bf{0.887(0.014)} & \bf{0.947(0.008)} & \bf{0.988(0.003)}\\
\hline
\end{tabular}
\end{center}
\end{table*}

\begin{table*}
\caption{Comparison of End-to-End face recognition systems on IJB-A}
\label{tbl:end_to_end}
\begin{center}
\begin{tabular}{|c|c|c|c|c|c|}
\hline
Face Detection & Face Alignment & Identity Descriptor & Metric Learning & Verif @FAR=0.01 & Ident Rank=1\\
\hline
DP2MFD~\cite{FD_BTAS2015} & LDDR~\cite{DBLP:journals/corr/KumarRPC16} & Chen~et~al.~\cite{chen2015end} & Joint Bayesian~\cite{chen2015end} & 0.776(0.033) & 0.834(0.017)\\
\hline
\multicolumn{2}{|c|}{HyperFace} & Sankaranarayanan~et~al~\cite{DBLP:journals/corr/Sankaranarayanan16a} & cosine & 0.871(0.01) & 0.925(0.01)\\
\hline
\multicolumn{2}{|c|}{HyperFace} & Sankaranarayanan~et~al~\cite{DBLP:journals/corr/Sankaranarayanan16a} & TPE~\cite{DBLP:journals/corr/Sankaranarayanan16a} & 0.90(0.01) & 0.932(0.01)\\
\hline
\multicolumn{2}{|c|}{HyperFace} & \textbf{Ours} & cosine & 0.889(0.01) & 0.939(0.01)\\
\hline
\multicolumn{3}{|c|}{\textbf{Ours}} & cosine & 0.893(0.01) & 0.941(0.008)\\
\hline
\multicolumn{3}{|c|}{\textbf{Ours}} & TPE~\cite{DBLP:journals/corr/Sankaranarayanan16a} & \bf{0.922(0.01)} & \bf{0.947(0.008)}\\
\hline
\end{tabular}
\end{center}
\end{table*}

For IJB-A dataset, given a template containing multiple faces, we generate a common vector representation by media pooling the individual face descriptors, as explained in~\cite{DBLP:journals/corr/Sankaranarayanan16a}. A naive way to measure the similarity  of a template pair, is by taking cosine distance between their descriptors. A better way is to learn an embedding space where features corresponding to similar pairs are close to each other while dissimilar pairs are far away. We train a Triplet Probabilistic Embedding (TPE)~\cite{DBLP:journals/corr/Sankaranarayanan16a} using the training splits provided by the dataset. Table~\ref{tbl:ijba} compares with recently published methods on IJB-A. We achieve state-of-the-results for the face identification task. Although we perform comparable to template-adaptaion learning (Crosswhite~et~al.~\cite{crosswhite2016template}) on verification task, we achieve a significantly faster query time  ($0.1s$  after
face detection per image pair). We get a consistent improvement of $2\%$ to $3\%$ over the baseline network~\cite{DBLP:journals/corr/Sankaranarayanan16a} for all metrics. 

We also compare our performance with end-to-end face recognition methods in Table~\ref{tbl:end_to_end}. Our method outperforms existing end-to-end systems which shows that training all the tasks in the pipeline simultaneously, reduces the error. We see a two-fold improvement, i.e., about $80\%$ performance gain is a result of improved identity descriptor and $20\%$ gain is due to improved face alignment.  

\subsection{Runtime}
We implemented our all-in-one network on a machine with 8 CPU cores and GTX TITAN-X GPU. It takes an average of $3.5$s to process an image. The major bottleneck for speed is the process of generating region proposals and passing each of them through the CNN. The second stage of our method takes merely $0.1$s of computation time.

\section{CONCLUSIONS AND FUTURE WORKS}
\label{sec:conclusion}
\begin{figure*}[htp!]
 \centering
\includegraphics[width=5.5cm, height=3.5cm]{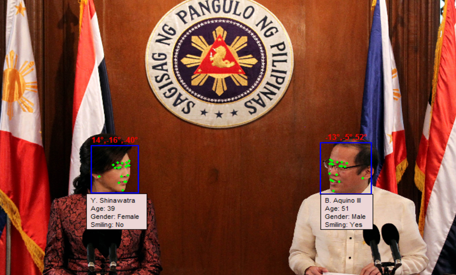}\hskip5pt\includegraphics[width=5.5cm,height=3.5cm]{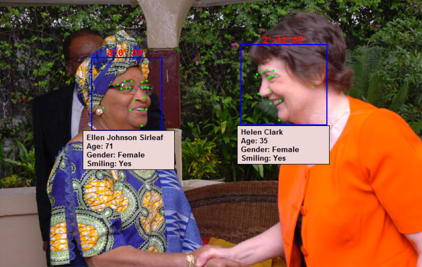}\hskip5pt\includegraphics[width=5.5cm,height=3.5cm]{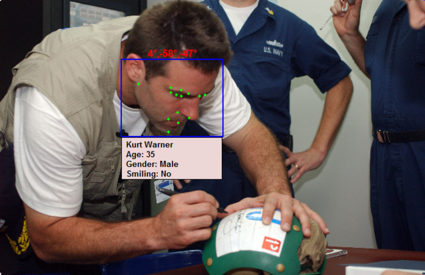}\\
\vskip5pt
\includegraphics[width=5.5cm, height=3.5cm]{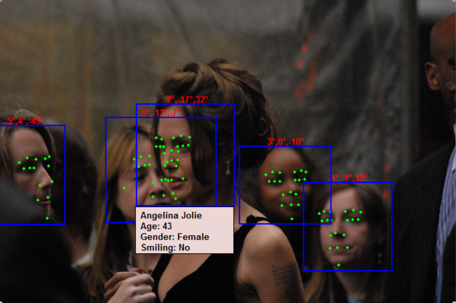}\hskip5pt\includegraphics[width=5.5cm,height=3.5cm]{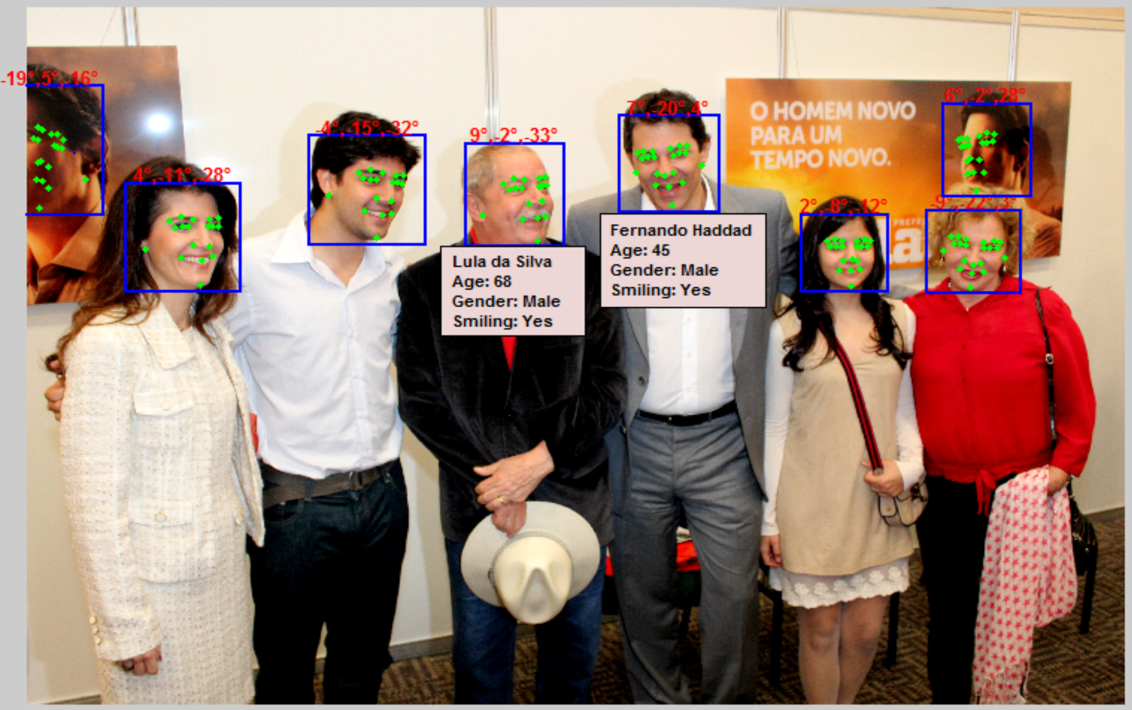}\hskip5pt\includegraphics[width=5.5cm,height=3.5cm]{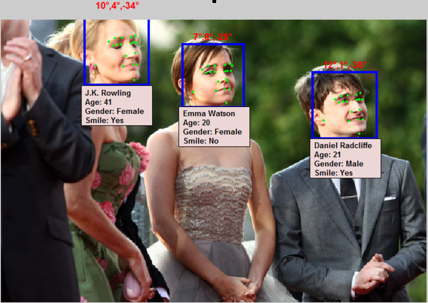}\\
\vskip5pt
\includegraphics[width=3.7cm, height=5.0cm]{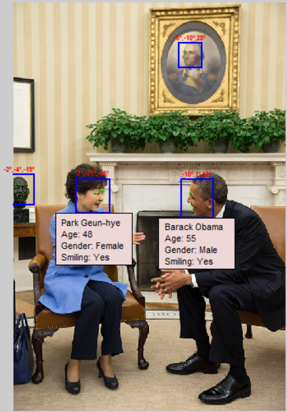}\hskip5pt\includegraphics[width=3.7cm,height=5.0cm]{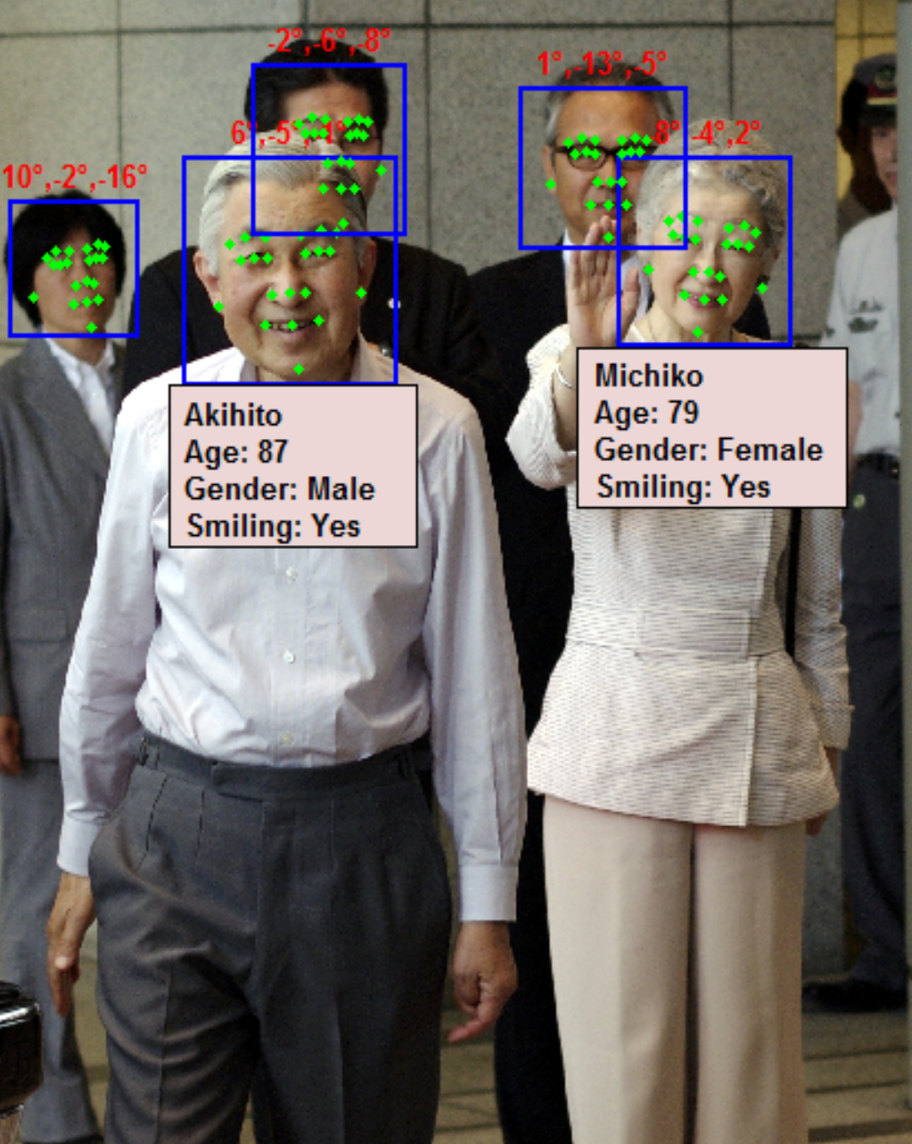}\hskip5pt\includegraphics[width=3.5cm,height=5.0cm]{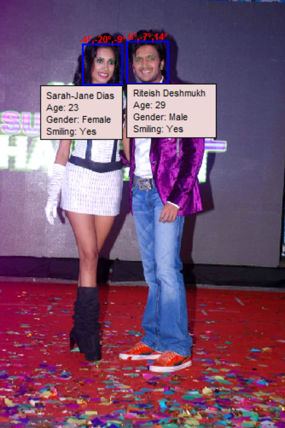}\\
\vskip5pt
\caption{Qualitative results of our method.  The blue boxes denote detected faces. The green dots provide the landmark locations. Pose estimates for each face are shown on top of the boxes in the order of roll, pitch and yaw. The predicted identity, age, gender and smile attributes are shown below the face-boxes. Although the algorithm generates these attributes for all the faces, we show them only for subjects that are present in the IJB-A dataset for better image clarity.}
\label{fig:quantative_results}
\end{figure*}

In this paper we presented a multi-task CNN-based method for simultaneous face detection, face alignment, pose estimation, gender and smile classification, age estimation and face verification and recognition. Extensive experiments on available unconstrained datasets show that we achieve state-of-the-art results for majority of these tasks. Our method performs significantly better than HyperFace, even though both of them use the MTL framework. This work demonstrates that subject-independent tasks benefit from domain-based regularization and network initialization from face recognition task. Also, the improvement in face verification and recognition performance compared to~\cite{DBLP:journals/corr/Sankaranarayanan16a} clearly suggests that MTL helps in learning robust feature descriptors. In future, we plan to extend this method for other face-related tasks and make the algorithm real time. Several qualitative results of our method are shown in Figure~\ref{fig:quantative_results}. 

\section{ACKNOWLEDGMENTS}

This research is based upon work supported by the Office of the Director of National Intelligence (ODNI), Intelligence Advanced Research Projects
Activity (IARPA), via IARPA R\&D Contract No. 2014-14071600012. The views and conclusions contained herein are those of the authors and should
not be interpreted as necessarily representing the official policies or endorsements, either expressed or implied, of the ODNI, IARPA, or the U.S. Government. The U.S. Government is authorized to reproduce and distribute reprints for Governmental purposes notwithstanding any copyright annotation
thereon.


%
%
%
%
%

{\small
\bibliographystyle{ieee}
\bibliography{UltraFace}
}

\end{document}